\documentclass[a4paper, 10pt, conference]{ieeeconf} 
\AtBeginDocument{\let\autocite\cite}

\IEEEoverridecommandlockouts                              

\usepackage{balance}
\usepackage{url}
\usepackage{cite}
\usepackage{graphicx} 
\usepackage{flushend}
\usepackage{booktabs}
\usepackage{tikz}
\usepackage{tabularx}

\newcommand\copyrighttext{%
  \footnotesize \textcopyright 2026 IEEE. Permission from IEEE must be obtained for all uses, in any current or future
  media, including reprinting/republishing this material for advertising or promotional
  purposes, creating new collective works, for resale or redistribution to servers or
  lists, or reuse of any copyrighted component of this work in other works.}
\newcommand\copyrightnotice{%
\begin{tikzpicture}[remember picture,overlay]
\node[anchor=south,yshift=10pt] at (current page.south) {\fbox{\parbox{\dimexpr\textwidth-\fboxsep-\fboxrule\relax}{\copyrighttext}}};
\end{tikzpicture}%
}

\begin{document}
\IEEEoverridecommandlockouts
\overrideIEEEmargins

\title{\LARGE \bf
  Toward User-Mediated Self-Repair in Ubiquitous Robots Through Goal-Oriented Agentic AI
}

\author{Morten Roed Frederiksen$^{1}$
  \thanks{{$^{1}$Morten Roed Frederiksen {\tt\small mrof@itu.dk} is affiliated with the Data Systems and Robotics (DSAR) Department of The IT-University of Copenhagen Denmark.}}
}

\maketitle
\copyrightnotice
\begin{abstract}
Ubiquitous robotic systems often lack traditional visual interfaces, necessitating resilient natural language interaction for maintenance and repair tasks. This paper presents a goal oriented agentic AI architecture designed to enable non-expert users to perform technical repairs through situated dialogue. The framework utilizes a multi-layered approach that decouples high-level strategic planning from reactive conversational execution to transform unconstrained human instructions into a structured hierarchy of goals. We conducted a study involving twenty participants to evaluate the system's efficacy using a physical hardware testbed. The architecture achieved a 95\% task completion rate, and participants reported positive self-efficacy following real-time guidance that adapted to conversational diversions and linguistic variations. A comparative analysis with an online baseline revealed that the transition to a physical environment significantly decreased perceived social presence (p=.0005), and trust and competence, ($p=.037$), while the agentic framework remained robust throughout the interaction. These findings indicate that goal oriented agentic AI can support the sustainability of body-worn technologies by empowering users to perform critical maintenance in ubiquitous contexts.
\end{abstract}

\section{Introduction}
The transition toward ubiquitous robotic systems worn directly on the human body introduces a novel set of environmental and operational challenges \cite{Zhao02102023, chen23}. While traditional robotics often operate in controlled industrial settings, wearable technologies are subjected to the dynamic mechanical strains of the human form, including constant movement, perspiration, and varied physical pressure \cite{ Zhao02102023, Heikenfeld2018WearableSM}. Because ubiquitous devices frequently suffer mechanical failure from sustained physical contact, there is an under-researched opportunity in empowering users to facilitate their own self-repairing strategies to ensure a sustainable future for these technologies.

However, the increasing complexity of robotic architectures presents a significant barrier to user-led maintenance. As robots become more complex, the logic required to repair them often undergoes a parallel increase in difficulty. This challenge is primarily rooted in the limits of human working memory, which can become overwhelmed when a system presents high intrinsic complexity \cite{Sweller1988CognitiveLD, Hancock07}. When technical features are extensive or internal dependencies are opaque, the user's ability to form a coherent mental model of the robot is undermined \cite{Kiesler2002MentalMO}. Instructional design for complex robotic systems has historically relied on static manuals or linear tutorials. These methods often fail in the context of self-repair as repair tasks frequently involve branching logic and internal dependencies that exceed the user's immediate cognitive capacity\cite{buttussi21, Henderson2011ExploringTB}.

\subsection{Agentic AI in a Ubiquitous Context}
This paper proposes the utilization of agentic artificial intelligence (AI) as a mediator to facilitate robot self-repair. By implementing a multi-layered agentic architecture, the robot can self-introduce its own repair requirements and guide the user through sophisticated maintenance procedures. This approach aims to foster an anthropomorphic interpretation that could positively shape the interactions with the robot \cite{Frederiksen2022RobotVA, Frederiksen2019AugmentingTA, Frederiksen2019ASC}. 
\begin{figure}[h]
\centering
\includegraphics[width=0.48\textwidth]{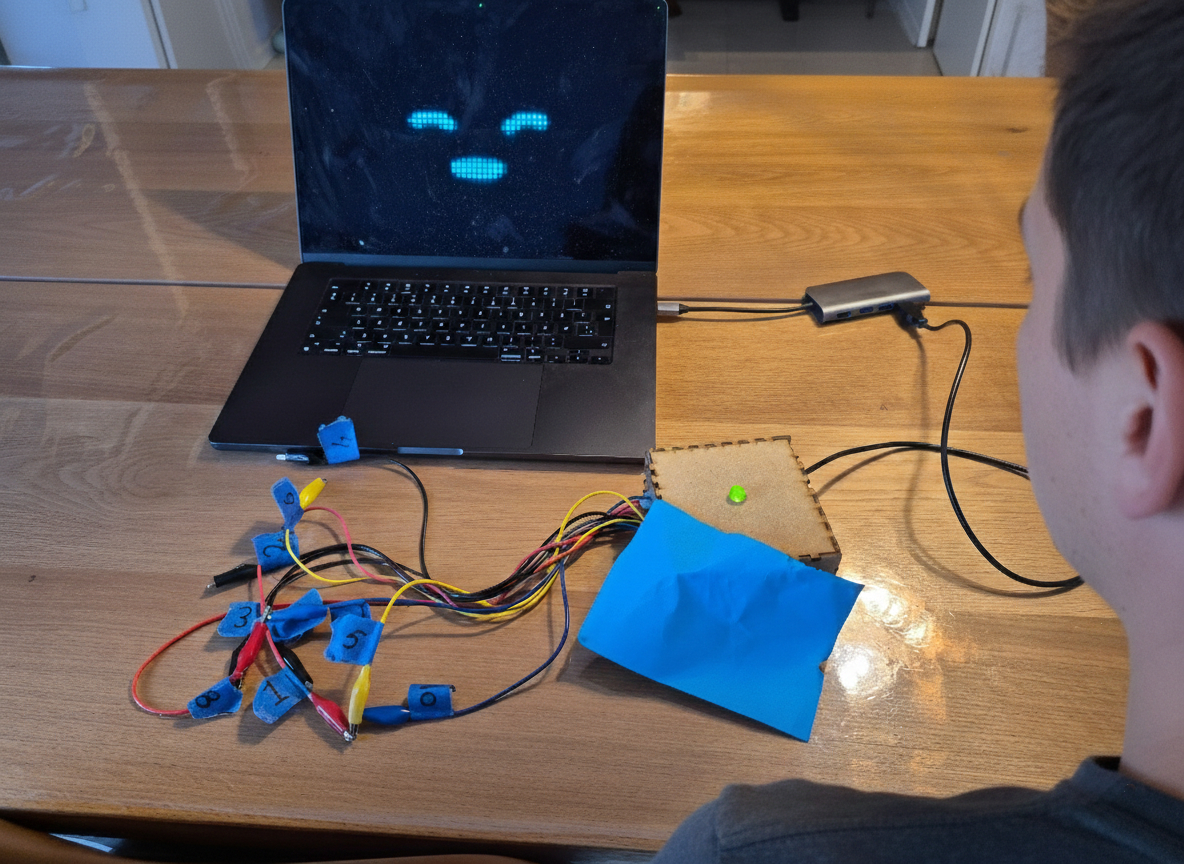}
\caption{Physical repair testbed used in the human-robot interaction experiment. Participants interacted with the animated robot face while repairing a hardware puzzle consisting of ten numbered wires, crocodile clips, a color-coded configuration sticker, and a confirmation lamp. The robot guided participants through a conditionally branched repair procedure in which the correct wire connections depended on whether the sticker was blue or green.}
\label{fig:test_setup}
\end{figure}
The inherent close proximity to the human body and the often hidden nature of ubiquitous robots (such as those operating within a pocket or under clothing) render them particularly well-suited for dialogue-based interfaces \cite{10.1145/191666.191732, Brewster2003MultimodalI}. Because these devices are frequently used in a concealed or eyes-free state to maintain discretion \cite{Profita2013DontMM, chulhong2015}, traditional visual-only instructional modalities are impractical \cite{Yi2012ExploringUM, Brewster2003MultimodalI, Zhu2025WearableIH}. Dialogue-based interaction allows for a private, hands-on repair process that preserves the user's focus on the tactile interaction while leveraging natural language to navigate complex diagnostic tasks \cite{Marge2020SpokenLI}. This alignment between form and interface transforms the device into a pocket-sized friend that can communicate its needs directly to the wearer \cite{Frederiksen2024TowardAP}. By framing the device as a social partner through anthropomorphic interaction \cite{Singh2023AnthropomorphismAS}, we intend to prime the user to view the repair process as a collaborative task. This approach aims to enrich the learning experience and improve the long-term sustainability of ubiquitous robotic systems \cite{Epley2007OnSH, Leite2013SocialRF, abanovi2018RobotsFU, Kim2015IntelligenceTF}.

\subsection{Open-Ended and Self-Repairing}
The proposed architecture departs from rigid troubleshooting manuals by adopting an open-ended pedagogical approach. Rather than forcing users through a fixed diagnostic tree, the system utilizes a strategic agent to monitor repair objectives and conversational context in real-time. This enables non-linear interaction where explanations are reframed based on user confusion or partial progress. Critically, the framework emphasizes temporal decoupling, allowing users to pause and resume repairs without penalty, thereby aligning the learning process with the realities of real-world maintenance.

To bridge the gap between system logic and human language, the framework employs an agent to handle task decomposition. This component transforms unconstrained human explanations into structured hierarchies of goals and subgoals, explicitly modeling internal dependencies and failure recovery paths. By delegating task management to this multi-layered agentic structure, the robot can self-introduce its repair requirements, effectively transforming a complex technical burden into a collaborative, social interaction that enhances both system longevity and user self-efficacy.

To test the system, we designed a human-robot interaction experiment, shown in Figure \ref{fig:test_setup}, in which participants collaborated with a conversational robot interface to complete a physical repair task. The physical testbed was designed as a controlled proxy for ubiquitous robot repair, capturing key properties such as diagnostic branching, spoken guidance, physical manipulation, and confirmation of repair success. This task came in different variations performed by test participants guided by and interacting with the goal oriented AI. The interaction required the AI to both elicit task-relevant information from participants and infer the participants’ current mental model and progress within the sequence of repair steps. The results of the experiment demonstrated that the agentic architecture facilitated a 95\% task completion rate, effectively navigating the complexities of real-world hardware maintenance. While a comparative analysis with a purely digital baseline indicated a significant decrease in perceived social presence and trust within the physical environment ($p<.05$), participants nevertheless reported high perceived helpfulness and post-interaction self-efficacy. These findings suggest that although the reality gap of physical repair introduces additional interactional and physical task demands, the goal-oriented approach remains robust, ensuring the functional sustainability of ubiquitous robotic systems through resilient, situated dialogue. 
The main contributions of this paper are threefold. First, we present a goal-oriented multi-agent architecture that separates pre-interaction goal decomposition, persistent state tracking, strategic goal management, and real-time conversational execution. Second, we demonstrate how this architecture can support open-ended, non-linear repair dialogue in a physical ubiquitous robotics scenario where users must complete a conditionally branched hardware task without relying on a traditional screen-based instructional interface. Third, we provide an empirical evaluation with a physical testbed and compare the results with a prior online baseline, showing that the system maintains high task completion and perceived helpfulness despite the additional demands of situated hardware interaction.

\section{Related Work}

This paper builds upon architectural frameworks that bridge high-level semantic reasoning with low-level robotic execution. We focus on two distinct yet converging domains: speech-driven interfaces for ubiquitous interaction and agentic task planning for embodied systems.

\subsection{Speech-Driven Interfaces in Ubiquitous Robotics}
Early research in Speech User Interfaces (SUI) prioritized command safety and navigational efficiency by enforcing rigid, finite-state grammars. Peltola et al. \cite{Peltola1999ADS} optimized handheld device interaction by dynamically restricting the active vocabulary to the current UI state, effectively eliminating out-of-domain errors at the cost of flexibility. As computing became ubiquitous, Goose et al. \cite{goose2003, Goose2003AugmentedRI} extended this to different contexts, utilizing audio-only augmented reality to guide maintenance tasks where visual attention was required on the physical machinery.

Contemporary approaches have moved beyond command-and-control to semantic comprehension. Marge et al. \cite{Marge2020SpokenLI} highlight the shift toward situated dialogue where robots resolve ambiguities through multi-turn conversation rather than explicit re-prompting. In the context of "eyes-free" wearable robotics, this shift is critical. While recent multimodal systems \cite{Liao2022RealityTalkRS} integrate gaze and gesture to resolve intent, our work addresses the specific constraint of concealed robotics, where visual joint attention is impossible. 

\subsection{Task-Oriented Agentic AI}
The transition from symbolic planning to agentic AI has been driven by the zero-shot reasoning capabilities of LLMs. Wei et al. \cite{Wei2022ChainOT} demonstrated that "Chain-of-Thought" (CoT) prompting allows models to decompose complex logical problems into intermediate steps, a prerequisite for sequential robotic repair.

Recent frameworks have formalized this decomposition into executable agentic loops. Yao et al. \cite{Yao2022ReActSR} introduced Reasoning and Acting, where agents interleave reasoning traces with environmental actions to update plans dynamically. In embodied contexts, this is exemplified by SayCan \cite{Ahn2022DoAI}, which grounds high-level natural language instructions into low-level primitives by scoring actions based on their physical feasibility (affordance functions) rather than just semantic probability.

However, generic planners often fail in specialized hardware maintenance due to a lack of domain constraints. Our approach diverges by implementing a bifurcated architecture similar to Voyager \cite{Wang2023VoyagerAO}, which separates a "slow-thinking" curriculum manager from a "fast-thinking" execution actor. While their system applies this to digital exploration, we adapt the dual-agent structure to physical hardware, utilizing the "slow" agent for structural logic and the "fast" agent for social fluency and latency reduction.

\section{Methods}

\subsection{System Architecture}
\begin{figure}[h]
\centering
\includegraphics[width=0.48\textwidth]{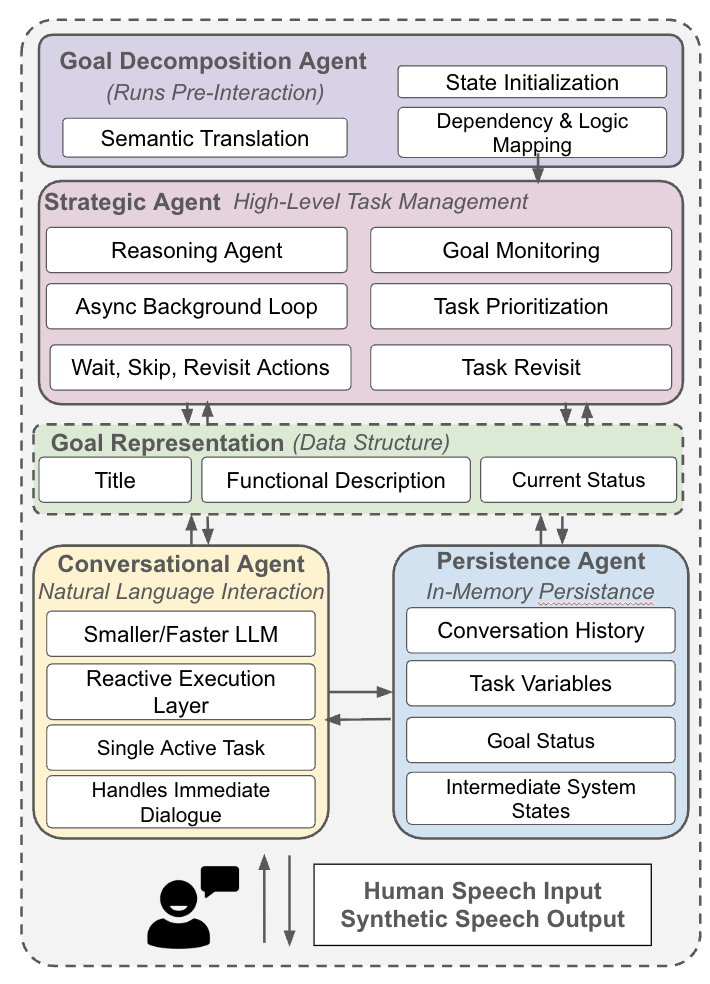}
\caption{Goal-oriented multi-agent architecture. Human speech is processed by the conversational agent, which produces real-time responses to the user while operating on the active goal selected by the strategic agent. The goal decomposition agent creates the initial goal hierarchy before interaction, while the persistence agent stores conversation history, task variables, goal status, and intermediate system states.}
\label{fig:agentic_architecture}
\end{figure}

We developed a goal-oriented agentic AI framework designed to facilitate the completion of complex, non-linear conversational repair tasks. The architecture is structured as a hierarchical multi-agent system (MAS) comprising four functional units: the \textit{Goal Decomposition Agent}, the \textit{Strategic Agent}, the \textit{Conversational Agent}, and the \textit{Persistence Handler Agent}. The system architecture is shown in Figure \ref{fig:agentic_architecture}. We evaluated the framework through open-ended interactions with 20 university participants who completed a physical robot repair task. This section describes the agentic AI architecture, the physical and online study procedures, the collected measures, and the comparative analysis.

\subsubsection{Goal Decomposition Agent}
The \textit{Goal Decomposition Agent} serves as the primary semantic bridge between unconstrained human instruction and robotic execution, transforming natural language into a rigorous hierarchy of goals and subgoals. This component operates as a "slow-thinking" processor, utilizing the \textit{Llama 3.1 70b} parameter model to interpret complex instructions and map internal dependencies. Unlike the conversational layer, this agent executes prior to the initiation of the live interaction, which removes the requirement for real-time inference speeds and allows the system to generate high-fidelity goal representations. For reproducibility, the agent was provided with the following verbatim instruction to generate the experimental goal structure: \textit{"I need you to create a series of goals to enable a robot to explain to a human how the human can repair it. Goal one is to find out which color sticker is on the box in front of the user. It can be either green or blue. The next three goals depend on the first answer. If it’s a blue sticker then have the user connect wires 1 and 3 and wires 2 and 4. If it’s a green sticker have the user connect wires 1 and 2 and wires 3 and 4. Finally have the user verify that the confirmation bulb light is showing."} By identifying these dependencies and potential branching logic during this pre-computation phase, the agent ensures the robot can navigate hardware variations or repair errors autonomously. This approach fosters an anthropomorphic interpretation  \cite{Roselli2025HowCM, Epley2007OnSH} that shapes the interaction into a social partnership, effectively enriching the learning experience while ensuring the system remains anchored to the necessary structural logic.
Each operational goal was defined by three distinct fields: a title, a functional description, and a current status. \textit{Goal Decomposition Agent} retained sole responsibility for managing goal sequences and conditional branching, as no explicit metadata regarding goal dependencies or temporal logic was injected during the pre-processing phase. Throughout the interaction, the \textit{persistent agent} maintained these goals as unformatted plaintext, providing a computationally lightweight state-tracking mechanism for the multi-agent system. This architectural separation ensures that high-level structural logic is derived dynamically from the system state rather than from static predefined scripts.

\subsubsection{The Persistence handler Agent}
The output of the decomposition agent is passed to the \textit{persistent agent}. This agent facilitates long-term interaction coherence by managing memory persistence throughout the human-robot engagement. To maintain a stable internal state while allowing for flexible, adaptive dialogue, the agent decouples memory management from both goal strategy and immediate conversational behavior. This architecture utilizes a \textit{ChromaDB} database \cite{chromadb_github} to store critical session data, including comprehensive conversation history, goal-specific variables provided by the user, and intermediate system states required for error recovery and conditional logic evaluation. Longitudinal coherence is often significant in ubiquitous or intermittent interaction scenarios, as it may enable the system to revisit prior instructions, prevent redundant explanations, and ensure consistent behavior across interaction turns \cite{Leite2013SocialRF, Bickmore2005EstablishingAM}. By functioning as the system's memory layer, the persistent agent can allow the robot to maintain an evolving mental model of the repair progress, reinforcing the user's perception of the device as a reliable social partner \cite{1291665}.

\subsection{Conversation handling agent}
The conversational agent is the user-facing component of the system. It serves as a high-throughput, reactive language-based interface for real-time natural language interaction within ubiquitous robotic systems, which are often physically concealed or lack a traditional visual interface. To ensure interaction fluency and minimize latency in environments with diminished visual feedback mechanisms, the agent uses the Qwen 30B large language model, balancing linguistic capacity with the responsiveness required for synchronous dialogue. Functioning as a reactive execution layer, the agent is constrained to the single active goal selected by the strategic agent, focusing on procedural explanation, user clarification, and validation of goal-satisfaction criteria.
Upon successful completion of a specific objective, the agent signals completion to the strategic planner but does not independently advance the overall workflow, ensuring that high-level structural logic remains the domain of the strategic layer.

\subsubsection{The Strategic Agent}
The \textit{strategic agent} governs high-level goal management and goal monitoring through a centralized goal board, which maintains the status of all discrete repair steps categorized as pending, active, completed, or skipped alongside their governing conditional logic. By synthesizing user input with the accumulated system state, the agent determines the active goal and enforces sequential or conditional progression in a coherent, goal-directed manner. Architecturally, this agent is powered by a high-parameter large language model (LLM) whose significant reasoning depth facilitates a comprehensive analysis of the longitudinal conversation history. To maintain the responsiveness required for ubiquitous interaction, the strategic agent operates within an asynchronous background execution loop, separating complex structural reasoning from the primary interaction flow. This meta-cognitive layer enables dynamic goal prioritization and the execution of pedagogical revisits if the system state indicates that a previously addressed objective requires further clarification. This approach ensures that the interaction remains aligned with predefined pedagogical requirements while facilitating the non-linear transitions and error recovery necessary for user-led maintenance. By framing the robot as a social partner capable of self-correcting its instructional path, the architecture enhances the user's self-efficacy and promotes an anthropomorphic interpretation that enriches the overall learning experience.

\subsection{Experimental Procedure and Testbed}

\subsubsection{Social Robotic Interface}

The robotic interface was implemented using the Python-based Pygame framework to facilitate, open-ended social interactions. To preserve a consistent frame rate and avoid latency-induced breaks in the interaction, the system used a distributed architecture: the graphical interface was hosted on a MacBook Pro, while large language model (LLM) inference was offloaded to a dedicated secondary server. 

\textit{System behavior and perception.} To project social agency, the interface exhibited autonomous naturalistic behaviors, including periodic blinking, smiling, and saccadic gaze shifts. These anthropomorphic features were designed to improve user trust and engagement. Multi-modal sensing governed the interaction state through the following mechanisms:

\begin{itemize}
    \item Visual Engagement: Real-time face detection via a high-definition camera served as the primary interaction trigger, with the robot initiating engagement only upon established eye contact.
    \item Auditory Processing: A continuous microphone feed monitored ambient sound pressure levels; when input exceeded a calibrated decibel threshold, the system engaged the OpenAI Whisper framework for speech-to-text (STT) transcription.
    \item Linguistic Sanitization: To ensure fluid vocalization by the text-to-speech (TTS) engine, all LLM-generated responses underwent automated sanitization to remove special characters and emojis before output.
\end{itemize}

\subsubsection{Physical study}

The study included 20 participants (N=20) recruited from the university, ranging in age from 18 to 44 years (12 male, 7 female, 1 non-binary). During the experiment, participants were presented with the robot face and a hardware testbed consisting of a wooden box equipped with a green status lamp. Ten numbered wires with crocodile clips hung from the side of the box, and a color-coded sticker (blue or green) was placed on the top to identify the repair task variation. To complete the repair, participants had to inform the conversational agent of the sticker color, obtain the specific wiring instructions for that configuration, and verify the activation of the status lamp. The internal circuitry ensured the lamp would only illuminate upon the correct connection of four specific wires. While the interaction remained open-ended and allowed for user diversion, the agentic AI was designed to steer the dialogue toward fulfilling the specific repair objectives.

Each participant proceeded through the following experimental protocol:

\begin{enumerate} 
\item Participants were provided with a pre-study questionnaire regarding age, gender, and familiarity with robotics and AI chatbots. 
\item This questionnaire included excerpted items from the Negative Attitude toward Robots Scale (NARS) and was administered via a tablet to minimize experimenter bias. \item Participants initially engaged in a two-minute casual conversation with the robot to become familiar with the interaction's turn-taking tempo. This phase was driven by the Qwen 30b model, which was prompted to be friendly and humorous. 
\item Participants then transitioned to the interaction with the agentic AI framework through the same robotic interface. 
\item The robot initiated the repair sequence by introducing the mechanical problem and requesting the user's assistance with its initial goal. 
\item Once the correct wires were connected and the user confirmed to the AI that the status lamp was active, the system concluded the task with a closing statement and the experimenter ended the session. 
\item The participant completed a post-experiment questionnaire evaluating the quality of the interaction and the system's pedagogical effectiveness. 
\end{enumerate}

Self-efficacy was assessed only after the interaction, meaning that this measure captures participants’ perceived capability following system use but does not establish a change from a pre-interaction baseline. Objective statistics for each interaction were stored alongside the questionnaire answers. A single set of the objective performance metrics was missing and excluded from the objective comparison. The rest of the data, including total turn counts, message lengths, and total session durations, were calculated for each interaction and merged with the Likert-scale responses for statistical analysis. To provide a comprehensive evaluation of the framework's efficacy across different modalities, we benchmark the current physical results against data from a prior online study \cite{frederiksen2026goal}. The online baseline used the same underlying goal-oriented agentic architecture and the same branching repair logic, but presented the task through a digital interface rather than a physical hardware testbed. Participants in the online study interacted with the system while completing a virtual version of the wiring task, in which selecting or connecting digital elements replaced the manipulation of physical wires and crocodile clips. Because the goal structure and task logic were held constant across the two studies, the comparison provides an initial indication of how situated physical interaction may affect user perception. This comparison provides a clearer view of how situatedness and physical embodiment influence user perception and task performance when interacting with the same goal-oriented agentic architecture.
\section{Results}
\subsection{Task Completion and Behavioral Outcomes}

Nineteen of the twenty participants successfully completed the puzzle task. Task completion was objectively verified through successful termination of the repair sequence and confirmation of the status lamp. Consequently, the physical study achieved a 95\% task completion rate, indicating that the agentic system reliably guided most users through the multi-step, conditionally branched task despite variations in interaction length.

\subsection{Perceived Helpfulness, Trust, and Social Presence}

Post-interaction evaluations indicated that participants generally perceived the AI as competent and trustworthy, though evaluations of social presence were more moderate. Composite scales were constructed from theoretically aligned questionnaire items. Both scales demonstrated acceptable to high internal consistency (Social Presence: Cronbach’s $\alpha = .80$; Trust and Competence: $\alpha = .93$).

Participants rated Trust and Competence well above the neutral midpoint of the 7-point Likert scale ($M = 4.87, SD = 1.57$), whereas Social Presence was rated closer to neutrality ($M = 4.09, SD = 1.45$).

To quantify the magnitude of these deviations from neutrality, standardized effect sizes were computed using Cohen’s $d$. As detailed in Table \ref{tab:psychometrics}, evaluations of Trust and Competence showed a medium effect size ($d=0.56$), while Social Presence showed a negligible effect size ($d=0.06$), suggesting that while users trusted the system's instructions, they did not strongly perceive it as a social partner.

\begin{table}[ht]
\centering
\caption{Participant Evaluations ($N=20$)}
\label{tab:psychometrics}
\scriptsize 
\begin{tabularx}{\columnwidth}{@{}l X X X X@{}} 
\toprule
Construct & $M$ & $SD$ & $\alpha$ & $d$ \\
\midrule
Trust \& Comp. & 4.87 & 1.57 & .93 & 0.56 \\
Social Presence & 4.09 & 1.45 & .80 & 0.06 \\
\bottomrule
\addlinespace
\multicolumn{5}{l}{\textit{Note.} $d$ is deviation from neutral midpoint (4.0).}
\end{tabularx}
\end{table}

\subsection{Relationship Between Prior AI Familiarity and Post-Interaction Trust}

To examine whether prior experience with AI influenced participants’ evaluations of the agent, a Pearson correlation was conducted between a Prior AI Familiarity index and the Trust and Competence composite. Results revealed no significant correlation, $r = -0.07, p = .775$.

This finding indicates that participants' trust in the system was independent of their prior familiarity with AI technologies. The evaluations of competence appear to stem primarily from the immediate interaction performance rather than because of pre-existing attitudes toward AI.

\subsection{Quantitative Analysis of Interaction Efficiency and Robustness}

To evaluate the operational efficiency of the agentic workflow, we compared participant interaction lengths against a theoretical optimal path of 9 conversational turns. Analysis of the interactions yielded a mean completion length of 19.7 turns ($SD=20.6$). A one-sample t-test indicates that the observed mean is statistically significantly higher than the theoretical optimum ($t(19) = 2.27, p = .035$).

The high standard deviation suggests substantial variability in user behavior, with several interactions extending beyond the necessary steps (e.g., due to chit-chat or repeated clarifications), while others adhered closely to the optimal path.

\begin{table*}[t] 
\centering
\caption{System Robustness and Recovery Against Non-Normative Inputs}
\label{tab:robustness}
\begin{tabularx}{0.8\textwidth}{l c X} 
\toprule
\textbf{Input Type} & \textbf{Frequency} & \textbf{Recovery Strategy} \\
\midrule
Code-switching & 10.5\% & Language mirroring (e.g., ``do you speak $<$language$>$?'') and intent processing. \\
\addlinespace
Meta-queries & 21.1\% & Refusal of out-of-scope requests (e.g., ``tell me a joke'') followed by task re-synchronization. \\
\addlinespace
Standard Flow & 68.4\% & Nominal state-management and goal-oriented flow. \\
\bottomrule
\end{tabularx}
\end{table*}

The system’s robustness was further illustrated by its capacity for rapid state recovery, as shown in Table \ref{tab:robustness}, when encountering non-normative inputs. Deviations were observed in the form of linguistic code-switching (10.5\% of sessions), where users switched to languages such as Danish (e.g., asking the robot if it speaks their native language), and divergent meta-queries (21.1\% of sessions). In the latter cases, users attempted to engage in social chit-chat (e.g., ``tell me a joke about whales'') or query the system's version. The agent consistently handled these by either briefly engaging before redirecting or politely refusing the request to refocus on the repair task.

\subsubsection{Baseline Methodology and Comparison}
To evaluate the efficacy of the agentic framework in a physical context, we utilize performance data from a previous iteration of the system tested in a purely digital environment as a comparative baseline. This online study involved $N=70$ participants and utilized a virtual representation of the wiring task. The inclusion of this baseline serves two primary functions: first, it establishes the fundamental capabilities of the agentic AI logic across a large sample size; second, it allows for a direct evaluation of how physical embodiment and the ``hidden'' nature of ubiquitous sensors impact user perception.

This comparison is relevant given that ubiquitous robots are often designed for discrete use \cite{Schwind19, OMETOV2021108074}. By framing the robot with anthropomorphic characteristics, we intend to prime the users to view the device as a social partner, thereby enriching the interaction quality \cite{Zotowski2014AnthropomorphismOA}. The physical testbed specifically evaluates whether the robot's goal-oriented persona remains effective when the AI must mediate tangible hardware repairs rather than digital puzzles. By maintaining identical goal structures between the online and physical versions, we can potentially view the effects of the situated environment on user self-efficacy.

\subsection{Comparative Baseline Analysis}
The results of the physical experiment ($N=20$) were compared against the $N=70$ online baseline to provide a comprehensive view of the system capabilities. These results are contextualized here against the situated hardware performance.

\begin{table}[h]
\centering
\caption{Comparative Metrics: Online Baseline vs. Physical Testbed}
\label{tab:comparison}
\begin{tabular}{lcc}
\toprule
\textbf{Metric} & \textbf{Online ($N=70$)} & \textbf{Physical ($N=20$)} \\ \midrule
Perceived Understanding & 4.21 & 4.25 \\
System Helpfulness & 4.21 & 4.60 \\
Conversational Turns ($M$) & 11.4 & 19.7 \\
Task Completion Rate & 100\% & 95\% \\
\bottomrule
\end{tabular}
\end{table}

As shown in Table \ref{tab:comparison}, while the Perceived Understanding remained stable between versions ($M=4.21$ vs. $M=4.25$), the physical implementation maintained a high rating in System Helpfulness ($M=4.60$ vs. $M=4.21$) in spite of the added complexity of the real-world repair task. This suggests that the agentic AI's ability to provide situated, real-time feedback is valued when the user is confronted with the tactile complexity of hardware repair. This aligns with findings that morphology and sensory configuration perform adequately when supported by an effective robot behavior \cite{Thomaz2008TeachableRU, Richter2025ImprovingHT, Chi2024InteractiveHT}.

\begin{table*}[t]
\centering
\caption{Statistical Comparison: Online Baseline vs. Physical Interaction}
\label{tab:stats_comparison}
\begin{tabular}{lccccc}
\toprule
\textbf{Metric} & \textbf{Online ($N=70$)} & \textbf{Physical ($N=20$)} & \textbf{Difference} & \textbf{$p$-value} & \textbf{Result} \\
\midrule
Social Presence & $M=5.55$ & $M=4.09$ & $-1.46$ & $.0005$ & \textbf{Significant Decrease} \\
Trust \& Competence & $M=5.74$ & $M=4.87$ & $-0.87$ & $.037$ & \textbf{Significant Decrease} \\
Conversational Turns & $M=11.4$ & $M=19.7$ & $+8.3$ & $.095$ & Not Significant (Trend) \\
System Helpfulness & $M=4.21$ & $M=4.60$ & $+0.39$ & $.37$ & Not Significant \\
Perceived Understanding & $M=4.21$ & $M=4.25$ & $+0.04$ & $.89$ & Not Significant \\
\bottomrule
\end{tabular}
\end{table*}

Despite the increase in average conversational turns (19.7 in the physical study compared to 11.4 in the baseline), participants reported relatively high Perceived Self-Efficacy following the physical task ($M = 4.95$). This result is consistent with the framework supporting users during the task, although the post-only measure does not demonstrate that the interaction increased self-efficacy. The higher turn count likely reflects the added interactional and physical demands of manipulating wires compared to clicking digital nodes, requiring more clarification steps. The significant drop in Social Presence and Trust, as detailed in Table \ref{tab:stats_comparison}, suggests that the physical task may have been experienced as more consequential or demanding than the purely digital online task, which may have contributed to lower user ratings. These results demonstrate that while a digital AI can convey logic, a situated agentic system acts as a supportive partner, which is essential for the sustainable maintenance of ubiquitous robotic technologies. Consistent with previous research, the anthropomorphic framing served to improve trust and likeability, even when the robot was used in a concealed state \cite{Visser2012TheWI, Epley2007OnSH, Ferstl2021HumanOR}.
\section{Discussion}

\subsection{Efficacy of Situated Ubiquitous Interactions}
The primary objective for a non-screen-based user interface is to facilitate interactions that are effective and goal-oriented. The 95\% task completion rate observed in this study demonstrates the significant potential of the proposed framework, as the agentic system successfully guided nearly all participants through a multi-step, conditionally branched repair task. However, a significant disparity exists between the theoretical optimal path of 9 turns and the observed mean interaction length ($M = 19.7, SD = 20.6$). This increase in interaction duration—representing a trend toward higher turn counts compared to the 11.4 turns of the online baseline—likely stems from the compounding latencies of speech recognition and the physical manipulation of crocodile connectors. Despite this temporal cost, the framework demonstrated high robustness by maintaining a near-total completion rate, confirming that the agentic architecture effectively mediates technical information in environments where traditional visual feedback is reduced.

\subsection{Perceptual Disparity: Virtual vs. Physical Environments}
A critical finding is that Social Presence ($M = 4.09$) and Trust and Competence ($M = 4.87$) were significantly lower in the physical task than in the online baseline, where Social Presence was $M = 5.55$, $p=.0005$, and Trust and Competence was $M = 5.74$, $p=.037$. This disparity suggests a "reality gap" between simulation and situated interaction. In the online version, connecting wires is a low-stakes activity involving digital drag-and-drop mechanics. Conversely, the physical testbed involved interlocking connectors and the perceived risk of live electrical current. For participants lacking electronics exposure, the physical task may have been daunting; even with safety assurances, the perceived consequence of a wiring error remains high in a tangible environment. The online version carried no such consequences, which likely enhanced the feeling of safety and artificially boosted the perceived competence of the AI. In contrast, physical interaction grounds the user's perception in the tangible difficulty of the task, leading to a more conservative evaluation of the agent’s capabilities, as evidenced by the medium effect size for Trust ($d = 0.56$) compared to the negligible effect size for Social Presence ($d = 0.06$).

\subsection{Goal-Oriented Safety and Limitations}
For ubiquitous robots lacking visual UI confirmation, maintaining a safe and aligned state is paramount. The system's robustness was validated by its capacity for rapid state recovery, even when encountering meta-queries (21.1\%) or linguistic code-switching (10.5\%). An interpretation of these results should however be done in lieu of their limitations. First, the physical evaluation involved a relatively small sample of twenty participants, which limits the statistical power and generalizability of the findings. The results should therefore be interpreted as evidence of feasibility rather than as a definitive evaluation of the architecture across repair contexts. In addition, self-efficacy was measured only after the interaction, without a pre-test or matched control condition. Consequently, the findings describe participants’ perceived self-efficacy following system use but do not support a causal claim that the system increased self-efficacy. Second, the physical repair task was intentionally constrained to a conditionally branched wiring procedure in order to maintain experimental control and participant safety. Although this task captured key properties of self-repair, including branching logic, user clarification, and physical manipulation, it did not represent complex maintenance involving interacting faults or substantial diagnostic uncertainty. Future work should evaluate the architecture with larger samples, controlled pre-post comparisons, longer maintenance procedures, multiple fault types, richer diagnostic uncertainty, and repeated repair sessions over time.

\section{Conclusion}

This paper presented a goal oriented agentic AI architecture designed to facilitate user-mediated maintenance of ubiquitous robotic systems. The framework decouples high level strategic planning from reactive conversational execution to guide non expert users through hardware repair procedures. By transforming unconstrained natural language into a structured hierarchy of goals, the system enables technical interventions through situated dialogue.

The framework was evaluated through a physical testbed experiment involving twenty participants. The results show that the system maintained high task success, while participants reported positive self-efficacy following real-time guidance that adapted to conversational diversions. Given the exploratory sample, post-only self-efficacy measure, and constrained repair procedure, these findings should be interpreted as initial evidence of feasibility rather than evidence of causal improvement or broad generalizability. While the transition from digital simulation to a physical environment introduced additional physical task demands and lowered social perception, the system remained robust against linguistic variations. These findings indicate that goal oriented agentic AI can support the sustainability of body worn technologies by enabling user led maintenance in ubiquitous contexts

\balance

\bibliography{bibliography}

@article{Peltola1999ADS,
  title={A dictionary-adaptive speech driven user interface for a distributed multimedia platform},
  author={Johannes Peltola and Johan Plomp and Tapio Sepp{\"a}nen},
  journal={Proceedings 25th EUROMICRO Conference. Informatics: Theory and Practice for the New Millennium},
  year={1999},
  volume={2},
  pages={326-332 vol.2}
}

@article{Zhao02102023,
author = {Zhehui Zhao and Haoran Fu and Ruitao Tang and Bocheng Zhang and Yunmin Chen and Jianqun Jiang},
title = {Failure mechanisms in flexible electronics},
journal = {International Journal of Smart and Nano Materials},
volume = {14},
number = {4},
pages = {510--565},
year = {2023},
publisher = {Taylor \& Francis},
doi = {10.1080/19475411.2023.2261775},
eprint = { 
    
        https://doi.org/10.1080/19475411.2023.2261775
}
}

@inproceedings{10.1145/191666.191732,
author = {Mynatt, Elizabeth D. and Weber, Gerhard},
title = {Nonvisual presentation of graphical user interfaces: contrasting two approaches},
year = {1994},
isbn = {0897916506},
publisher = {Association for Computing Machinery},
address = {New York, NY, USA},
doi = {10.1145/191666.191732},
booktitle = {Proceedings of the SIGCHI Conference on Human Factors in Computing Systems},
pages = {166–172},
numpages = {7},
location = {Boston, Massachusetts, USA},
series = {CHI '94}
}

@article{Sweller1988CognitiveLD,
  title={Cognitive Load During Problem Solving: Effects on Learning},
  author={John Sweller},
  journal={Cogn. Sci.},
  year={1988},
  volume={12},
  pages={257-285}
}

@inproceedings{chulhong2015,
author = {Deterding, Sebastian and Lucero, Andrés and Holopainen, Jussi and Min, Chulhong and Cheok, Adrian and Waern, Annika and Walz, Steffen},
year = {2015},
month = {04},
pages = {2365-2368},
title = {Embarrassing Interactions},
doi = {10.1145/2702613.2702647}
}

@article{Zotowski2014AnthropomorphismOA,
  title={Anthropomorphism: Opportunities and Challenges in Human–Robot Interaction},
  author={Jakub Złotowski and Diane Proudfoot and Kumar Yogeeswaran and Christoph Bartneck},
  journal={International Journal of Social Robotics},
  year={2014},
  volume={7},
  pages={347 - 360}
}

@article{Visser2012TheWI,
  title={The World is not Enough: Trust in Cognitive Agents},
  author={Ewart de Visser and Frank Krueger and Patrick E. McKnight and Steven Scheid and Melissa A. B. Smith and Stephanie Chalk and Raja Parasuraman},
  journal={Proceedings of the Human Factors and Ergonomics Society Annual Meeting},
  year={2012},
  volume={56},
  pages={263 - 267}
}

@inproceedings{frederiksen2026goal,
  author    = {Frederiksen, Morten Roed},
  title     = {A Goal-Oriented Agentic Framework For Collaborative Branching Human-Robot Interactions},
  booktitle = {Proceedings of the 2026 IEEE Workshop on Advanced Robotics and its Social Impacts (ARSO)},
  year      = {2026},
  note      = {Under review}
}

@article{Chi2024InteractiveHT,
  title={Interactive Human-Robot Teaching Recovers and Builds Trust, Even With Imperfect Learners},
  author={Vivienne Bihe Chi and Bertram F. Malle},
  journal={2024 19th ACM/IEEE International Conference on Human-Robot Interaction (HRI)},
  year={2024},
  pages={127-136}
}

@article{Richter2025ImprovingHT,
  title={Improving Human-Robot Teaching by Quantifying and Reducing Mental Model Mismatch},
  author={Phillip Richter and Heiko Wersing and Anna-Lisa Vollmer},
  journal={ArXiv},
  year={2025},
  volume={abs/2501.04755}
}

@article{Thomaz2008TeachableRU,
  title={Teachable robots: Understanding human teaching behavior to build more effective robot learners},
  author={Andrea Lockerd Thomaz and Cynthia Lynn Breazeal},
  journal={Artif. Intell.},
  year={2008},
  volume={172},
  pages={716-737}
}

@article{OMETOV2021108074,
title = {A Survey on Wearable Technology: History, State-of-the-Art and Current Challenges},
journal = {Computer Networks},
volume = {193},
pages = {108074},
year = {2021},
issn = {1389-1286},
doi = {https://doi.org/10.1016/j.comnet.2021.108074},
author = {Aleksandr Ometov and Viktoriia Shubina and Lucie Klus and Justyna Skibińska and Salwa Saafi and Pavel Pascacio and Laura Flueratoru and Darwin Quezada Gaibor and Nadezhda Chukhno and Olga Chukhno and Asad Ali and Asma Channa and Ekaterina Svertoka and Waleed Bin Qaim and Raúl Casanova-Marqués and Sylvia Holcer and Joaquín Torres-Sospedra and Sven Casteleyn and Giuseppe Ruggeri and Giuseppe Araniti and Radim Burget and Jiri Hosek and Elena Simona Lohan}
}

@article{Yi2012ExploringUM,
  title={Exploring user motivations for eyes-free interaction on mobile devices},
  author={Bo Yi and Xiang Cao and Morten Fjeld and Shengdong Zhao},
  journal={Proceedings of the SIGCHI Conference on Human Factors in Computing Systems},
  year={2012}
}

@inproceedings{Schwind19,
author = {Schwind, Valentin and Deierlein, Niklas and Poguntke, Romina and Henze, Niels},
year = {2019},
month = {02},
pages = {},
title = {Understanding the Social Acceptability of Mobile Devices using the Stereotype Content Model},
doi = {10.1145/3290605.3300591}
}

@ARTICLE{1291665,
  author={Breazeal, C.},
  journal={IEEE Transactions on Systems, Man, and Cybernetics, Part C (Applications and Reviews)}, 
  title={Social interactions in HRI: the robot view}, 
  year={2004},
  volume={34},
  number={2},
  pages={181-186},
  doi={10.1109/TSMCC.2004.826268}}

@article{Bickmore2005EstablishingAM,
  title={Establishing and maintaining long-term human-computer relationships},
  author={Timothy W. Bickmore and Rosalind W. Picard},
  journal={ACM Trans. Comput. Hum. Interact.},
  year={2005},
  volume={12},
  pages={293-327}
}

@misc{chromadb_github,
  author = {Jeff Huber and Anton Troynikov},
  title = {Chroma},
  year = {2023},
  publisher = {GitHub},
  journal = {GitHub repository},
  howpublished = {\url{https://github.com/chroma-core/chroma}}
}

@article{Frederiksen2019ASC,
  title={A Systematic Comparison of Affective Robot Expression Modalities},
  author={Morten Roed Frederiksen and Kasper Støy},
  journal={2019 IEEE/RSJ International Conference on Intelligent Robots and Systems (IROS)},
  year={2019},
  pages={1385-1392}
}

@article{Frederiksen2019AugmentingTA,
  title={Augmenting the audio-based expression modality of a non-affective robot},
  author={Morten Roed Frederiksen and Kasper St{\o}y},
  journal={2019 8th International Conference on Affective Computing and Intelligent Interaction (ACII)},
  year={2019},
  pages={144-149}
}

@article{Frederiksen2022RobotVA,
  title={Robot Vulnerability and the Elicitation of User Empathy},
  author={Morten Roed Frederiksen and Katrin Fischer and Maja J. Matari{\'c}},
  journal={2022 31st IEEE International Conference on Robot and Human Interactive Communication (RO-MAN)},
  year={2022},
  pages={52-58}
}

@article{Ferstl2021HumanOR,
  title={Human or Robot?: Investigating voice, appearance and gesture motion realism of conversational social agents},
  author={Ylva Ferstl and Sean C. Thomas and C{\'e}dric Guiard and Cathy Ennis and Rachel Mcdonnell},
  journal={Proceedings of the 21st ACM International Conference on Intelligent Virtual Agents},
  year={2021}
}

@article{Leite2013SocialRF,
  title={Social Robots for Long-Term Interaction: A Survey},
  author={Iolanda Leite and Carlos Martinho and Ana Paiva},
  journal={International Journal of Social Robotics},
  year={2013},
  volume={5},
  pages={291 - 308}
}

@article{Roselli2025HowCM,
  title={How culture modulates anthropomorphism in Human-Robot Interaction: A review.},
  author={Cecilia Roselli and Leonardo Lapomarda and Edoardo Datteri},
  journal={Acta psychologica},
  year={2025},
  volume={255},
  pages={
          104871
        }
}

@inproceedings{Kim2015IntelligenceTF,
  title={Intelligence Technology for Ubiquitous Robots},
  author={Jong-Hwan Kim and Sheir Afgen Zaheer and Si-Jung Ryu},
  booktitle={Intelligent Assistive Robots},
  year={2015}
}

@article{abanovi2018RobotsFU,
  title={Robots For Us: Organizational and Community Perspectives on the Collaborative Design of Ubiquitous Robots},
  author={Selma Sabanovic},
  journal={Proceedings of the 31st Annual ACM Symposium on User Interface Software and Technology},
  year={2018}
}

@article{Epley2007OnSH,
  title={On seeing human: a three-factor theory of anthropomorphism.},
  author={Nicholas Epley and Adam Waytz and John T. Cacioppo},
  journal={Psychological review},
  year={2007},
  volume={114 4},
  pages={
          864-86
        }
}

@article{Singh2023AnthropomorphismAS,
  title={Anthropomorphism and Social Robotics in Kazuo Ishiguro's Klara and the Sun (2021)},
  author={Aman Deep Singh},
  journal={2023 IEEE 11th Region 10 Humanitarian Technology Conference (R10-HTC)},
  year={2023},
  pages={420-427}
}

@article{Frederiksen2024TowardAP,
  title={Toward Anxiety-Reducing Pocket Robots for Children},
  author={Morten Roed Frederiksen and Kasper St{\o}y and Maja J. Mataric},
  journal={2024 21st International Conference on Ubiquitous Robots (UR)},
  year={2024},
  pages={13-20}
}

@article{Marge2020SpokenLI,
  title={Spoken Language Interaction with Robots: Research Issues and Recommendations, Report from the NSF Future Directions Workshop},
  author={Matthew Marge and Carol Y. Espy-Wilson and Nigel G. Ward and Abeer Alwan and Yoav Artzi and Mohit Bansal and Gil and Blankenship and Joyce Yue Chai and Hal Daum{\'e} and Debadeepta Dey and Mary P. Harper and Thomas M. Howard and Casey and Kennington and Ivana Kruijff-Korbayov{\'a} and Dinesh Manocha and Cynthia Matuszek and Ross Mead and Raymond and Mooney and Roger K. Moore and Marilyn Ostendorf and Heather Pon-Barry and Alex Rudnicky and Matthias and Scheutz and Robert St. Amant and Tong Sun and Stefanie Tellex and David R. Traum and Zhou Yu},
  journal={Comput. Speech Lang.},
  year={2020},
  volume={71},
  pages={101255}
}

@article{Zhu2025WearableIH,
  title={Wearable Intelligent Human–Machine Interfaces Ready for Sustainable Edge Computing Systems},
  author={Minglu Zhu and Shuhan He and Tao Chen and Chengkuo Lee},
  journal={AI Sensors},
  year={2025}
}

@inproceedings{Brewster2003MultimodalI,
  title={Multimodal 'eyes-free' interaction techniques for wearable devices},
  author={Stephen Anthony Brewster and Joanna Lumsden and Marek Bell and Malcolm Hall and Stuart Tasker},
  booktitle={International Conference on Human Factors in Computing Systems},
  year={2003}
}

@inbook{Hancock07,
author = {Hancock, Peter and Szalma, James},
year = {2007},
month = {01},
pages = {195-206},
title = {Stress and Neuroergonomics},
isbn = {9780195177619},
journal = {Neuroergonomics: the Brain at Work},
doi = {10.1093/acprof:oso/9780195177619.003.0013}
}

@inproceedings{Profita2013DontMM,
  title={Don't mind me touching my wrist: a case study of interacting with on-body technology in public},
  author={Halley P. Profita and James Clawson and Scott M. Gilliland and Clint Zeagler and Thad Starner and Jim Budd and Ellen Yi-Luen Do},
  booktitle={International Semantic Web Conference},
  year={2013}
}

@article{Heikenfeld2018WearableSM,
  title={Wearable sensors: modalities, challenges, and prospects.},
  author={Jason C. Heikenfeld and Andrew J. Jajack and John A. Rogers and Philipp Gutruf and Limei Tian and Tingrui Pan and Ronald A. Li and Michelle Khine and Jayoung Kim and Joseph Wang},
  journal={Lab on a chip},
  year={2018},
  volume={18 2},
  pages={
          217-248
        }
}

@article{Henderson2011ExploringTB,
  title={Exploring the Benefits of Augmented Reality Documentation for Maintenance and Repair},
  author={Steven J. Henderson and Steven K. Feiner},
  journal={IEEE Transactions on Visualization and Computer Graphics},
  year={2011},
  volume={17},
  pages={1355-1368}
}

@article{buttussi21,
author = {Buttussi, Fabio and Chittaro, Luca},
year = {2021},
month = {02},
pages = {1-15},
title = {A Comparison of Procedural Safety Training in Three Conditions: Virtual Reality Headset, Smartphone, and Printed Materials},
volume = {14},
journal = {IEEE Transactions on Learning Technologies},
doi = {10.1109/TLT.2020.3033766}
}

@article{Kiesler2002MentalMO,
  title={Mental models of robotic assistants},
  author={Sara B. Kiesler and Jennifer Goetz},
  journal={CHI '02 Extended Abstracts on Human Factors in Computing Systems},
  year={2002}
}

@article{chen23,
author = {Chen, Lufeng and Xie, Hongqin and Liu, Zicheng and Li, Bin and Cheng, Hong},
year = {2023},
month = {12},
pages = {},
title = {Exploring Challenges and Opportunities of Wearable Robots: A Comprehensive Review of Design, Human-Robot Interaction and Control Strategy},
volume = {12},
journal = {APSIPA Transactions on Signal and Information Processing},
doi = {10.1561/116.00000156}
}

@article{Yao2022ReActSR,
  title={ReAct: Synergizing Reasoning and Acting in Language Models},
  author={Shunyu Yao and Jeffrey Zhao and Dian Yu and Nan Du and Izhak Shafran and Karthik Narasimhan and Yuan Cao},
  journal={ArXiv},
  year={2022},
  volume={abs/2210.03629}
}

@article{Wang2023VoyagerAO,
  title={Voyager: An Open-Ended Embodied Agent with Large Language Models},
  author={Guanzhi Wang and Yuqi Xie and Yunfan Jiang and Ajay Mandlekar and Chaowei Xiao and Yuke Zhu and Linxi (Jim) Fan and Anima Anandkumar},
  journal={ArXiv},
  year={2023},
  volume={abs/2305.16291}
}

@inproceedings{Ahn2022DoAI,
  title={Do As I Can, Not As I Say: Grounding Language in Robotic Affordances},
  author={Michael Ahn and Anthony Brohan and Noah Brown and Yevgen Chebotar and Omar Cortes and Byron David and Chelsea Finn and Keerthana Gopalakrishnan and Karol Hausman and Alexander Herzog and Daniel Ho and Jasmine Hsu and Julian Ibarz and Brian Ichter and Alex Irpan and Eric Jang and Rosario M Jauregui Ruano and Kyle Jeffrey and Sally Jesmonth and Nikhil Jayant Joshi and Ryan C. Julian and Dmitry Kalashnikov and Yuheng Kuang and Kuang-Huei Lee and Sergey Levine and Yao Lu and Linda Luu and Carolina Parada and Peter Pastor and Jornell Quiambao and Kanishka Rao and Jarek Rettinghouse and Diego M Reyes and Pierre Sermanet and Nicolas Sievers and Clayton Tan and Alexander Toshev and Vincent Vanhoucke and F. Xia and Ted Xiao and Peng Xu and Sichun Xu and Mengyuan Yan},
  booktitle={Conference on Robot Learning},
  year={2022}
}

@article{Wei2022ChainOT,
  title={Chain of Thought Prompting Elicits Reasoning in Large Language Models},
  author={Jason Wei and Xuezhi Wang and Dale Schuurmans and Maarten Bosma and Ed H. Chi and F. Xia and Quoc Le and Denny Zhou},
  journal={ArXiv},
  year={2022},
  volume={abs/2201.11903}
}

@article{Liao2022RealityTalkRS,
  title={RealityTalk: Real-Time Speech-Driven Augmented Presentation for AR Live Storytelling},
  author={Jian Liao and Adnan Karim and Shivesh Singh Jadon and Rubaiat Habib Kazi and Ryo Suzuki},
  journal={Proceedings of the 35th Annual ACM Symposium on User Interface Software and Technology},
  year={2022}
}

@inproceedings{Goose2003AugmentedRI,
  title={Augmented Reality in the Palm of your Hand : A PDA-Based Framework Offering a Location-based , 3 D and Speech-Driven User Interface},
  author={Stuart Goose},
  year={2003}
}

@ARTICLE{goose2003,
  author={Goose, S. and Sudarsky, S. and Xiang Zhang and Navab, N.},
  journal={IEEE Pervasive Computing}, 
  title={Speech-enabled augmented reality supporting mobile industrial maintenance}, 
  year={2003},
  volume={2},
  number={1},
  pages={65-70},
  doi={10.1109/MPRV.2003.1186727}
  }
\bibliographystyle{IEEEtran}

\end{document}